\documentclass[runningheads]{llncs}
\usepackage{nicefrac}
\usepackage{amssymb}
\usepackage{amsmath}
\usepackage{booktabs}
\usepackage{graphicx}
\usepackage{color}
\usepackage{multirow}
\usepackage{nicefrac}
\usepackage{url}
\begin{document}
\title{TopNet: Topology Preserving Metric Learning for Vessel Tree Reconstruction and Labelling}
\titlerunning{TopNet}
\author{Deepak Keshwani\inst{1} \and
Yoshiro Kitamura\inst{1} \and
Satoshi Ihara\inst{1} \and
Satoshi Iizuka\inst{2} \and
Edgar Simo-Serra\inst{3}
}

% index{Keshwani, Deepak}
% index{Kitamura, Yoshiro}
% index{Ihara, Satosh}
% index{Iizuka, Satoshi}
% index{Simo-Serra, Edgar}

\authorrunning{D. Keshwani et al.}
\institute{Imaging technology center, Fujifilm corporation, Japan \and
Center for Artificial Intelligence Research, University of Tsukuba, Japan \and
Department of Computer science and engineering, Waseda university, Japan 
\email{deepak.keshwani@fujifilm.com}
}
\maketitle              % typeset the header of the contribution
\begin{abstract}
Reconstructing Portal Vein and Hepatic Vein trees from contrast enhanced abdominal CT scans is a prerequisite for preoperative liver surgery simulation. Existing deep learning based methods treat vascular tree reconstruction as a semantic segmentation problem. However, vessels such as hepatic and portal vein look very similar locally and need to be traced to their source for robust label assignment. Therefore, semantic segmentation by looking at local 3D patch results in noisy misclassifications. To tackle this, we propose a novel multi-task deep learning architecture for vessel tree reconstruction. The network architecture simultaneously solves the task of detecting voxels on vascular centerlines (i.e. nodes) and estimates connectivity between center-voxels (edges) in the tree structure to be reconstructed.  Further, we propose a novel connectivity metric which considers both inter-class distance and intra-class topological distance between center-voxel pairs. Vascular trees are reconstructed starting from the vessel source using the learned connectivity metric using the shortest path tree algorithm. A thorough evaluation on public IRCAD dataset shows that the proposed method considerably outperforms existing semantic segmentation based methods. To the best of our knowledge, this is the first deep learning based approach which learns multi-label tree structure connectivity from images.

\end{abstract}

\section{Introduction}
Primary liver cancer is the third most common cause of cancer mortality~\cite{center2011international}. Liver resection is currently the most prevalent treatment method with 5-year survival rates of up to 40\%\ \cite{kazaryan2010laparoscopic}. To perform a safe liver resection, preoperative surgical simulations have been found to be very useful~\cite{wakabayashi2015recommendations,mise2018has}. In these simulations, a 3D model of the liver which shows the anatomical structures of portal and hepatic veins together with tumour location is reconstructed. Using vascular reconstructions, blood flow patterns around the tumour is analysed to compute the resection region. The analysis of blood flow requires not just segmentation of vascular structures but their representation as a tree structure. 

\begin{figure}[t]
\centering
\includegraphics[width=\linewidth]{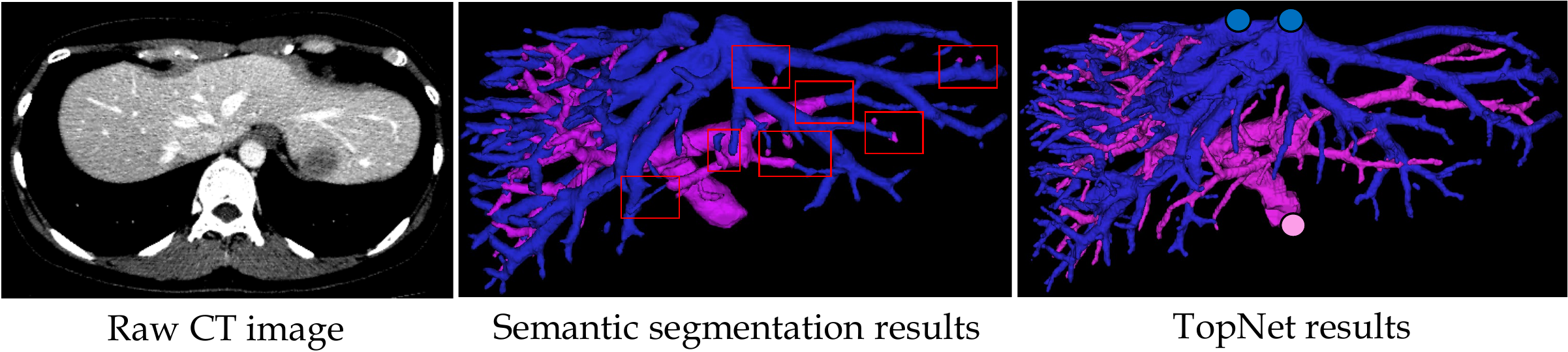}
\caption{Portal and hepatic vein structures obtained with semantic segmentation and proposed TopNet. The red rectangles indicate vessel misclassifications.} \label{fig:intro}
\end{figure}

Rule based methods for hepatic and portal vein reconstruction have been studied and a comprehensive literature survey can be found in~\cite{fraz2012blood}. Most popular rule based methods are based on global optimization techniques like graph cuts~\cite{zeng2017liver,bauer2010segmentation}. Graph cut based techniques use a handcrafted energy term between vascular nodes. A downside of such methods is that they work well only in conditions for which the rules were handcrafted. For instance when both portal and hepatic veins are high in contrast~\cite{zeng2017liver}. 
Recently, deep learning-based semantic segmentation methods have been proposed for liver vessel recognition tasks~\cite{ibragimov2017combining,kitrungrotsakul2017robust,kitrungrotsakul2019vesselnet}. In these methods, semantic segmentation is used to classify a voxel by extracting information from a local 3D patch (also known as the network’s receptive field) centered around it. However, since portal and hepatic veins look very similar locally, these methods result in noisy vessel misclassifications as shown in Figure~\ref{fig:intro}. Such noisy misclassifications magnify the tree reconstruction error. Additionally, within semantic segmentation literature, simultaneous segmentation of portal and hepatic vein from CT volumes has not been tackled. Either only hepatic~\cite{kitrungrotsakul2017robust,kitrungrotsakul2019vesselnet} or only portal vein~\cite{ibragimov2017combining} segmentation has been targeted. 

In this work, we propose TopNet, a deep metric learning method for vascular tree reconstruction. The main idea is to learn the connectivity metric between vascular voxel pairs rather than to assign absolute labels. Conceptually, TopNet is  a two stage network. The first stage computes center-voxels along the centerlines of all the vessels. The second stage learns the connectivity metric between center-voxel pairs in the form of topological distance between the voxels along the vascular tree. To generate a global tree, center-voxels are connected consecutively starting from the vessel source using the learned topology metric. We summarize the main contributions of this work as follows:

\begin{enumerate}
\item Proposed a novel multi task architecture designed for tree reconstruction, which detects nodes (voxels on vascular center lines) and estimates edges (connectivity between center-voxels) in the tree to be reconstructed. 
\item Proposed a novel topology metric which learns both inter-class distance and intra-class topological distance between vascular voxel pairs in multiple trees.
\item Verified that the proposed method achieves higher accuracy than existing methods with a large margin for portal and hepatic vein reconstruction.
\end{enumerate}

\section{Related work}
TopNet uses deep metric learning to connect detected center-voxels. In this regard, our approach is closer to deep metric learning approaches for image instance segmentation tasks ~\cite{de2017semantic,kong2018recurrent,fathi2017semantic,payer2018instance}. These methods map pixels from image space to features space such that pixels belonging to the same instance are close to each other and separated by a margin if they belong to different instances. The methods differ in how they measure similarity in the feature space. Among the similarity metrics, cosine similarity has been used extensively~\cite{kong2018recurrent,payer2018instance}. Another way in which these methods differ is the way pixels are clustered in the learned feature space to generate instances. In this work, the task is not just instance segmentation (vessel segmentation) but to learn connectivity within the instance (vessel trees). For this purpose, we propose a novel deep metric learning approach.

\section{TopNet}

\begin{figure}[t]
\centering
\includegraphics[width=\linewidth]{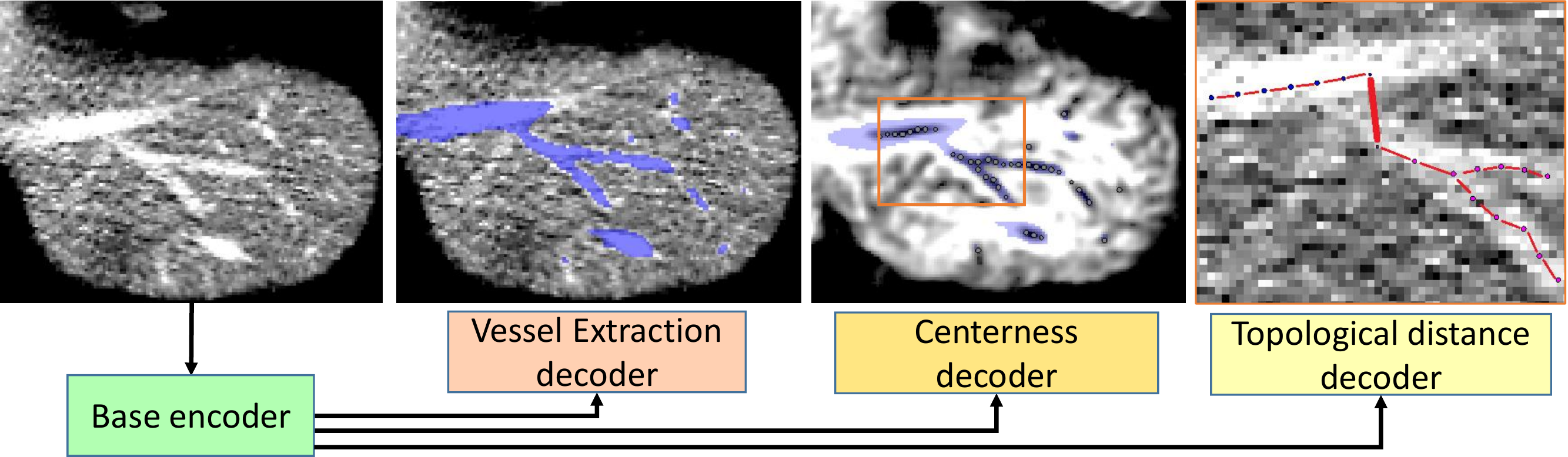}
\caption{TopNet methodology with base encoder and three task specific decoders.} \label{fig:TopNet}
\end{figure}

The proposed TopNet is a multi-task 3D Fully Convolutional Neural Network (3D-FCN).  All the tasks share a common base encoder and task specific decoders as shown in Figure~\ref{fig:TopNet}. The proposed architecture is similar to 3D-UNet~\cite{cciccek20163d}, with the difference that our approach has three decoders for three different tasks instead of one. Such Multi-task architectures with shared encoder and task specific decoders have also been proposed earlier~\cite{keshwani2018computation}. For detailed architecture, please refer to the supplementary material. The first task is extraction of all the vessels from the image. The second task is assigning a centerness score to all the voxels within the vessel mask such that the centermost voxel is assigned a minimum value. Using this centerness score, center-voxels are computed using non-maximum suppression. The final task is learning connectivity between center-voxels, which is proportional to topological distance between the center-voxels. Figure~\ref{fig:TopNet} shows the conceptual representation of topological distance with solid red lines. The network is trained by minimizing the sum of loss terms from each of the three tasks. The formulation of loss for each task is described in the following sections.

\subsection{Vessel Extraction Decoder}
The task of separating all the vessels from the background is formulated as a semantic segmentation problem. The loss function used for this task is the dice loss function which has shown to improve the accuracy in segmentation tasks with severe class imbalance~\cite{milletari2016v}. We compute vessel extraction loss $V_{loss}$ as

\begin{equation}
\label{eq:a}
	V_{loss}=1-\frac{2\sum_{i=1}^{N} {v}_{i}{v'}_{i}}{\sum_{i=1}^{N} {v}_{i} + \sum_{i=1}^{N} {v'}_{i}},
\end{equation}
where the sums run over the N voxels, of the ground truth binary vessel volume ${v}_{i} \in V$, and predicted vessel volume ${v'}_{i} \in V'$.

\subsection{Centerness Decoder}
The task of generating a centerness score map is formulated as a regression problem. The ground truth centerness score map is the distance transform of binary volume ${m}_{i} \in M$ representing the vessel center voxels. We consider the centerness score map only within the ground truth binary vessel volume $V$. The voxels outside the vessel region are masked out when computing the training loss. The training loss for the regression task is formulated as weighted $smooth_{L1}$ loss between the ground truth centerness score $S_i$ and predicted centerness score $S_i'$ summed over the voxels belonging to vessel mask $V$. We use $smooth_{L1}$ function as defined in~\cite{girshick2015fast} and the weight is inverse square of true centerness score i.e. $S_i^{2}$. Mathematically, the centerness loss $C_{loss}$  is given as
\begin{equation}
\label{eq:a}
	C_{loss}=\frac{1}{\sum_{i=1}^{N} {v}_{i}}\sum_{i\mid v_i=1} \frac{1}{S_i^2}\left(smooth_{L1}\right(S_i -S_i') ).
\end{equation}
%\begin{equation}
%\label{eq:b}
%smooth_{L1}\left(x\right)=\begin{cases}0.5(\sigma x) ^{2}& if \left | x \right | <1/\sigma ^{2} \\\ \left | x \right |-0.5/\sigma ^{2} & otherwise.\end{cases}
%\end{equation}
Weighting the loss function is essential because the loss would otherwise be highly biased towards the voxels away from the center.   During the inference, the centerness score is masked using the predicted vessel mask. The vessel center-points are extracted by first thresholding the centerness score map followed by applying non-maximum suppression (NMS) on negative centerness score map. In this work, we use a threshold of 1.5 and NMS window of $5\times5\times5$. 

\subsection{Topology distance Decoder}
Topological distance decoder outputs an 8-dimensional feature vector for each voxel in CT volume. The network is trained to map vessel center-voxels into the feature space such that the $L_2$ norm between two center-voxels in the feature space is proportional to the topological distance between along the vascular trees. For voxel pairs belonging to different vascular trees, the  $L_2$ norm is greater than a set margin. The pairwise loss between the two voxels $i$ and $j$ with associated vessel labels $l_{i}$ and $l_{j}$ is defined as
\begin{equation}
\label{eq:b}
	Top(x_i,x_j)=\begin{cases}smooth_{L1}\left(\parallel x_i-x_j\parallel -\alpha D_{ij} \right) & \forall i,j\mid l_{i}=l_{j} ,i\neq j, \\\gamma*\max(0,\left[K-\parallel x_{i}-x_{j}\parallel\right]) & otherwise.\end{cases}
\end{equation}

\noindent Here, $\parallel x_i-x_j\parallel$ is the $L_2$ norm between the feature representations of the two voxels, $D_{ij}$ is the topological distance between the voxels along the vascular tree. $\alpha$ and $\gamma$ are constants whose values are explained subsequently. To compute the topological loss, all such center voxel pairs which are in the local neighborhood of each other are considered. The total loss then becomes 

\begin{equation}
\label{eq:b}
T_{loss}=\frac{1}{n}\sum_{i\mid m_i=1}\\\ \sum_{j\in N_i}Top \left({x_i,x_j}\right).
\end{equation}

Here, $N_i$ is the set of voxels in the local neighborhood of voxel $i$ and $n$ is the total number of such voxel pairs for which the loss is computed. In this work, we use all voxel pairs which lie inside a 3D sphere of radius 15 voxels in the image space. Thus, the topological distance is normalized from 0 to 1 using the proportionality factor $\alpha = \nicefrac{1}{15}$. This roughly means that the voxels belonging to the same vascular tree will follow  $0\leq\parallel x_i-x_j\parallel\leq1$ and if they belong to different vascular trees, then $ \parallel x_i-x_j\parallel \geq3$. To balance out the two loss terms, we use $\gamma = \nicefrac{1}{3}$. 

\subsection{Vascular tree reconstruction}
Vascular tree is constructed by aggregating the center-voxels starting from the vessel sources using the topological distance metric learned by the TopNet. We train a separate network which outputs the location of portal and hepatic vein sources (see supplementary material). Typical portal (pink disc) and hepatic vein (blue disc) sources are shown in Figure~\ref{fig:intro}. For computed vessel sources, the vascular trees are constructed using Dijkstra’s multi-source shortest path tree algorithm. To that end, the undirected graph is formulated as follows. The vertices of the graph are center-voxels. The weighted edges of the graph are given by the set ${\left\{(\nicefrac{w_{ij}}{\alpha})^{2}\right\}}_{w_{ij}\leq2}$. Here, $w_{ij}=\parallel x_{i}-x_{j}\parallel$, $i\mid m_{i}=1$ and $j\mid m_{j}=1$  represents two center-voxels in a local neighborhood defined by a 3D sphere of radius 15. The term $(\nicefrac{w_{ij}}{\alpha})^{2}$ is essentially the square of topological distance between the two center-voxels $i$ and $j$. All the pairs for which $w_{ij\geq2}$ are not considered as feasible edges, since such edges most likely belong to different vascular trees.

\section{Experiments}
We divide the experiments into comparison and ablation studies. The comparison of TopNet is made against single task 3D-UNet~\cite{cciccek20163d} with the dice loss function~\cite{milletari2016v} baseline. This is a standard and competitive baseline for 3D image segmentation and an upgrade over existing 2D-FCN based existing liver vessel segmentation methods  ~\cite{ibragimov2017combining,kitrungrotsakul2017robust,kitrungrotsakul2019vesselnet}. In the ablation study, we investigate the performance with different learning metrics (dice loss based classification, cosine, and topology) using the same multi-task architecture. For the Multi-task dice loss based classification approach, the last two layers of topology distance decoder are replaced to output a two-channel probability map for each hepatic and portal vein. Here, we use a 2 class dice loss summed over all the center-voxels. Multi-task cosine metric learning method uses the exact same architecture as TopNet but uses cosine metric following from similar works~\cite{kong2018recurrent,payer2018instance}. The network uses cosine metric loss between voxel pairs $i$ and $j$ associated with feature vectors $x_{i}$ and $x_{j}$ and vessel labels $l_{i}$ and $l_{j}$ is defined as 

\begin{equation}
\label{eq:b}
L\left(x_i,x_j\right)=\begin{cases}1-0.5(1 + S_{ij})& \forall i,j\mid l_{i}=l_{j}, i\neq j,  \\\ 0.5(1 + S_{ij}) & otherwise.\end{cases}
\end{equation}
where, $S_{ij}=\frac{\parallel x_{i}^{T}x_j\parallel}{\parallel x_i\parallel\parallel x_j\parallel}\ $ is the cosine similarity metric. To reconstruct vascular tree using the learned metric as mentioned in section 3.4, the weighted edges of the graph are given by the set $ \left \{ E_{ij}(1-S_{ij}) \right \}_{S_{ij}\geq 0} $ Here, $E_{ij}$ is the Euclidean distance between the center-voxels.

%smooth_{L1}\left(x\right)=\begin{cases}0.5(\sigma x) ^{2}& if \left | x \right | <1/\sigma ^{2} \\\ \left | x \right |-0.5/\sigma ^{2} & otherwise.\end{cases}

\subsection{Dataset and preprocessing}
We used an internal dataset for training and both the internal as well as public IRCAD dataset for evaluation~\cite{IRCAD}. The characteristics of both the datasets is shown in Table~\ref{tab1}. For training, the CT values were normalized from 0 to 1, voxel spacing in all three dimensions was normalized to 0.7mm. We also crop the CT volume using the liver region before setting it as input to the network.

\begin{table}[t]
\caption{Characteristics of INTERNAL and public dataset IRCAD.}\label{tab1}
\centering
\setlength{\tabcolsep}{6pt}
\begin{tabular}{lllll}
\toprule
Dataset  & Training       & Test          & Contrast phase & Slice Thickness    \\
\midrule
INTERNAL & 115 CT volumes & 20 CT volumes & Late portal    & 0.5 $\sim$ 1.0 mm \\
IRCAD~\cite{IRCAD}    & None           & 20 CT volumes & Late portal    & 1.0 $\sim$ 4.0 mm \\
\bottomrule
\end{tabular}
\end{table}

\begin{figure}[]
\centering
\includegraphics[width=\linewidth]{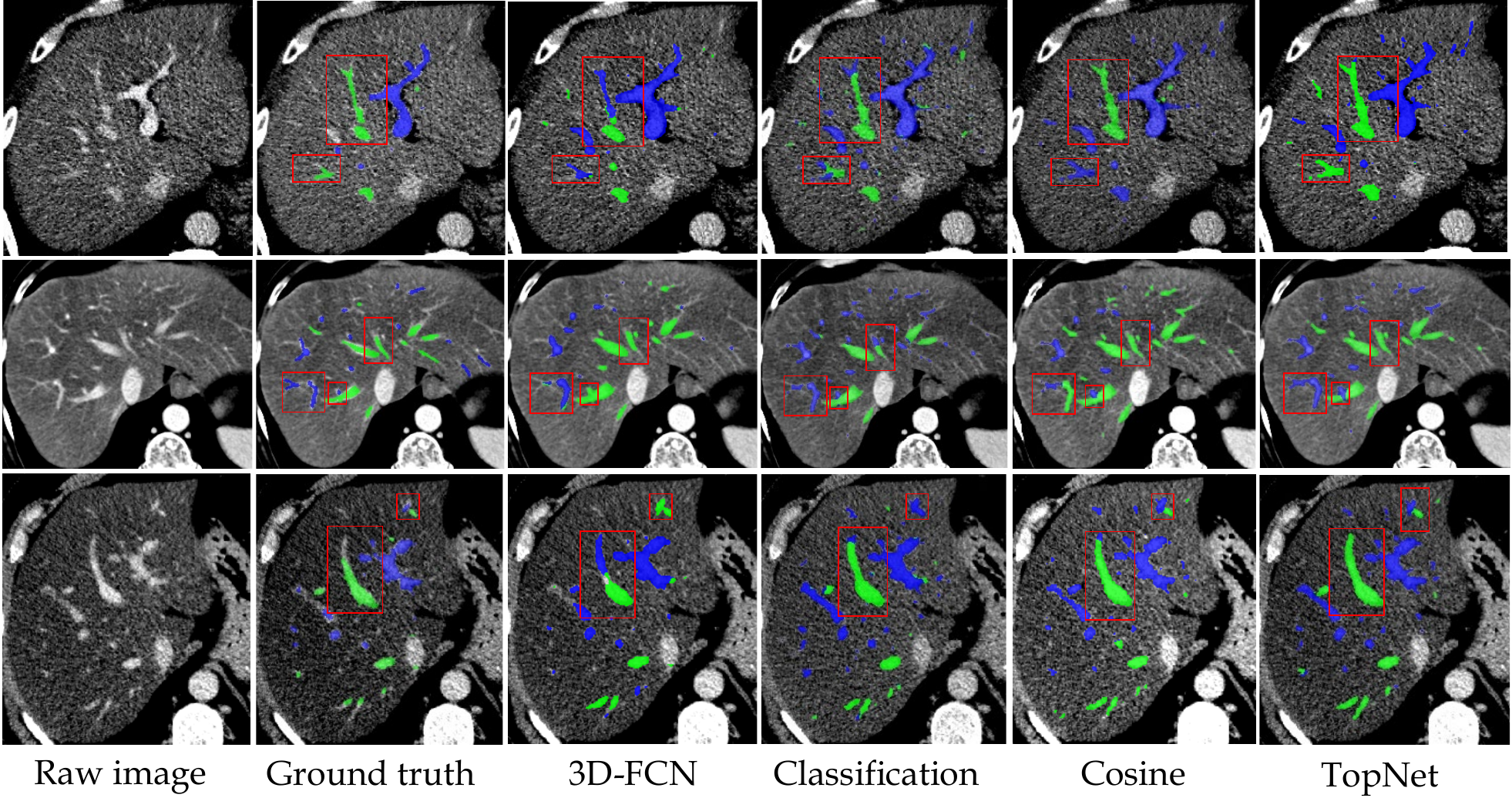}
\caption{Comparison between 3D-UNet and proposed TopNet for hepatic and portal vein extraction.  Hepatic vein is shown in green while portal vein is shown in blue. Red boxes show regions of interest for comparing misclassifications.} \label{fig:Results1}
\end{figure}

\subsection{Evaluation metrics and results}
We evaluate the results of portal and hepatic vein reconstruction by comparing respective ground truth and predicted centerlines. When comparing the centerlines, an exact overlap is not possible. For this, we introduce a variable tolerance term $\delta$ which is equal to the radius defined at a ground truth centerpoint following from~\cite{gegundez2011function}. Based on this, an illustration of True Positives (TP), False Positives (FP), False Negatives (FN), and True Negatives (TN) of portal vessel class is shown in the supplementary material. Using these metrics, we compute the dice coefficient, sensitivity and specificity for hepatic and portal centerlines.  Predicted centerpoints for which there are no associated ground truth centerpoints within the tolerance are ignored while computing the metrics. Usually, these points would be counted as false positives but we ignore them because in the IRCAD dataset, small vessels are often unlabelled.

\begin{figure}[t]
\centering
\includegraphics[width=\linewidth]{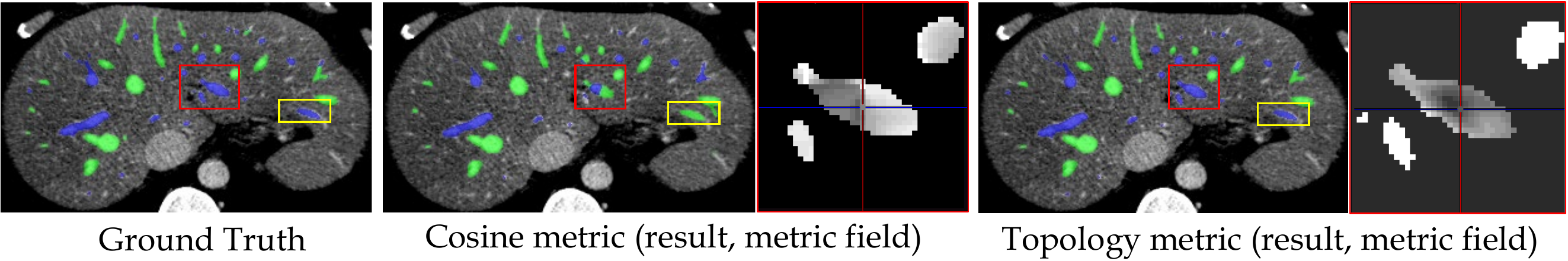}
\caption{Comparison between Topology and cosine metric learning.  Hepatic vein is shown in green while portal vein is shown in blue. Boxes show regions of interest for comparing misclassifications.} \label{fig:Results2}
\end{figure}

The proposed TopNet considerably outperforms the baseline i.e. single task 3D-UNet method with over 7-8\%\ improvement in all metrics as shown in Table~\ref{tab2}. When Multi-task dice loss classification is compared to single-task 3D-UNet which also uses the dice loss, a performance improvement of 3-4\%\ is seen. By this, we can say that multi-stage vessel segmentation works better than single-stage method. Within the Multi-task network architecture, TopNet gives the best results. We observe that if not for topology metric learning, the problem of abrupt change in vessel labels along the vascular tree (large and medium red boxes) persists as shown in Figure~\ref{fig:Results1}. Such misclassifications also occur when portal and hepatic veins cross each other in close vicinity (small red boxes in second and third row). We reason that such misclassifications arise because conventional margin based techniques like semantic segmentation or even cosine metric learning classify or seperate the center-voxels based on frequently seen vessel patterns in different liver regions within the training dataset. If unique vessel patterns are observed in test data, the classification fails. With Topology metric learning on the other hand, we explicitly constrain the network to learn to trace the vessels. To verify this direction of reasoning, we show the similarity metric field from a reference center-voxel to it's neighborhood voxels in a misclassified region in Figure~\ref{fig:Results1}. It can be seen that the Topology metric shows a smooth transitioning of the similarity field as compared to cosine metric which shows an abrupt change. Since the cosine metric learning method does not learn absolute labels, such misclassifications propagate as shown with the yellow box in Figure ~\ref{fig:Results2}. This explains why the cosine metric learning works poorly as compared to the multi-task classification approach as shown in Table~\ref{tab2}. 

\begin{table}[t]
\caption{Comparison between TopNet and existing methods.}\label{tab2}
\centering
\setlength{\tabcolsep}{2.9pt}
\begin{tabular}{llcccccc}
\toprule
\multirow{1}{*}{Method}          & \multirow{1}{*}{Vessel} & \multicolumn{3}{l}{\multirow{1}{*}{IRCAD}}                                                       & \multicolumn{3}{l}{\multirow{1}{*}{INTERNAL}}                                                   \\
\midrule
                                 &                         & Dice                           & Specificity                    & Sensitivity                    & Dice                           & Specificity                    & Sensitivity                    \\ 
\midrule
\textbf{Single Task}                      &                         &                                &                                &                                                     & \multicolumn{1}{l}{}                               &                                &                                \\ 
3D-UNet                   & Portal                  & 0.84                           & 0.85                           & 0.82                           & 0.88                           & 0.86                          & 0.87                            \\
\cite{cciccek20163d}                         & Hepatic                 & 0.84                           & 0.82                            & 0.85                           & 0.86                           & 0.87                            & 0.86                           \\ 
\midrule
\textbf{Multi Task}                      &                         &                                &                                &                                                     & \multicolumn{1}{l}{}                               &                                &                                \\ 
\multirow{1}{*}{Classification}                      & Portal                  & 0.87                            & 0.85                            & 0.88                           & 0.92                            & 0.93                            & 0.90                           \\
          & Hepatic                 & 0.86                           & 0.88                           & 0.85                           & 0.91                           & 0.90                           & 0.93                           \\ \cline{1-8}
\multirow{1}{*}{Cosine metric}               & Portal                  & 0.85                           & 0.85                           & 0.81                            & 0.88                           & 0.84                           & 0.9                            \\
                        & Hepatic                 & 0.80                          & 0.82                           & 0.86                           & 0.85                           & 0.90                           & 0.84                           \\ \hline
\multirow{1}{*}{TopNet} & Portal                  & {}\textbf{0.91} & {}\textbf{0.96}  & {}\textbf{0.91} & {}\textbf{0.95} & {}\textbf{0.96}  & {}\textbf{0.94} \\
                              & Hepatic                 & {}\textbf{0.92} & {}\textbf{0.91} & {}\textbf{0.96} & {}\textbf{0.94} & {}\textbf{0.94} & {}\textbf{0.96} \\ 
\bottomrule
\end{tabular}
\end{table}

\section{Conclusion}
We proposed a deep metric learning method which learns multi-label tree structure connectivity from images. We converted a hard problem of assigning absolute labels to vessels to a simpler one. Topological metric learning is simpler because it just learns local connectivity and does not require global image context. For global connectivity, we use the shortest path tree algorithm which uses the learned metric to connect vascular voxels. The results show that using topological metric learning, the issue of noisy misclassifications is resolved. We believe that the approach is general enough to be applicable to vessel segmentation tasks in other organs. This is because the network is constrained to learn connectivity rather than other absolute labels.  There are a few downsides to this approach as well. First, the method has multiple manually fine tuned parameters involved both in detecting the center-voxels and creating vascular trees using the learned topology metric. Second, the shortest path tree algorithm which is used to create global connectivity is sensitive to misdetection of vascular center voxels. Due to this, in the future it would be interesting to explore a method which can embed tree generation algorithms like the shortest path into the training process. 

%Hepato-Biliary-Pancreatic Surgery Division, The University of Tokyo, Japan\\ 

%
% the environments 'definition', 'lemma', 'proposition', 'corollary',
% 'remark', and 'example' are defined in the LLNCS documentclass as well.
%

\bibliographystyle{splncs04}
\bibliography{bibliography603}

\section*{Acknowledgements} 
This research was done in cooperation with the Hepato-Biliary-Pancreatic Surgery Division, the University of Tokyo Hospital. 
We would like to thank Prof. Kiyoshi Hasegawa, Dr. Junichi Kaneko, Dr. Ryugen Takahashi, and Dr. Yusuke Kazami for their valuable advice and curating a liver vessel dataset

\end{document}

\begin{figure}[]
\centering
\includegraphics[width=\linewidth]{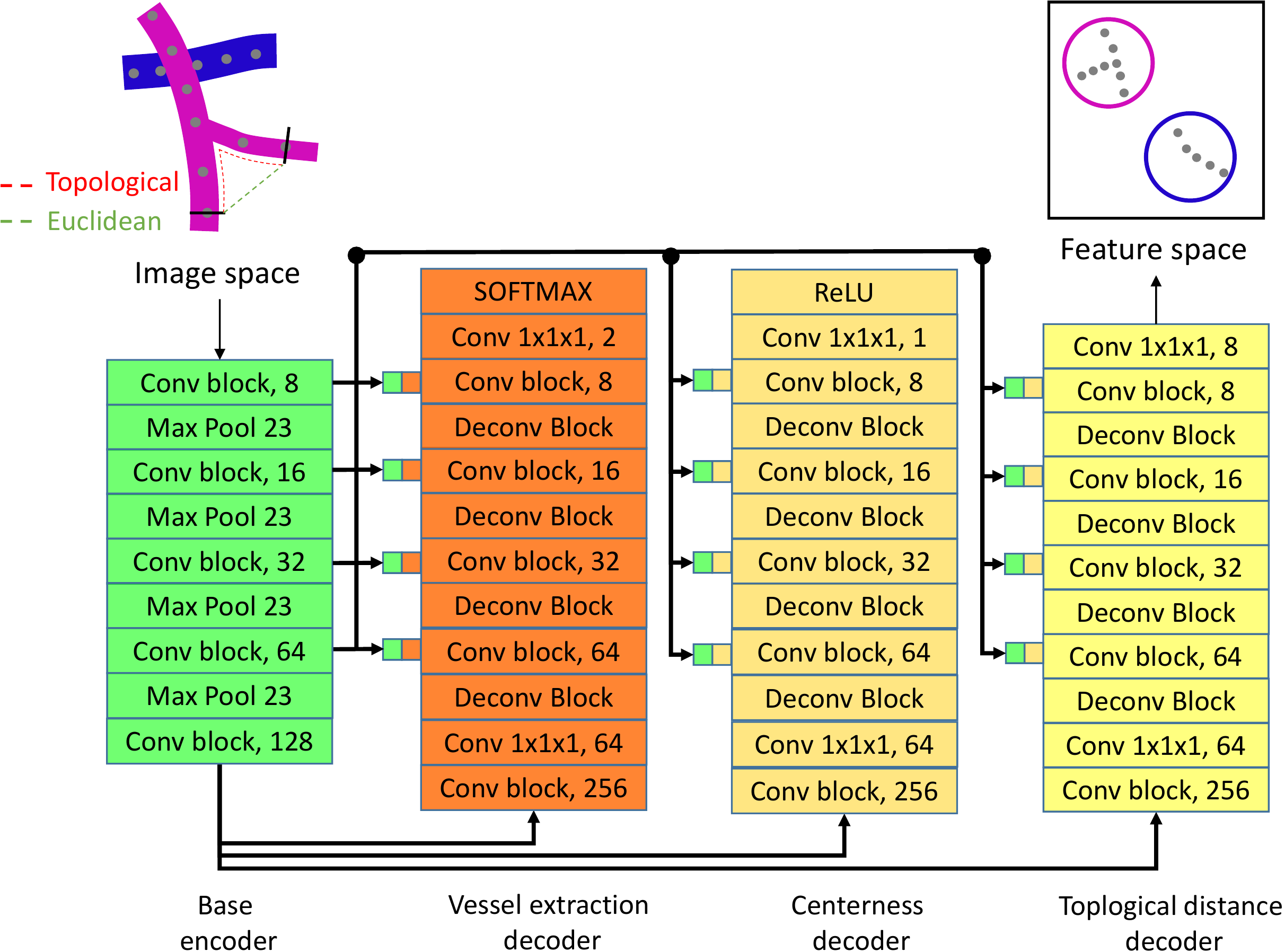}
\caption{TopNet Architecture: The Conv block in each encoder and decoder denotes two sets of $Conv[3\times3\times3] \rightarrow Batch Norm\rightarrow ReLU$ operations. The Deconv blocks are transposed convolutional operators with filter size $4\times4\times4$. Skip connections are used between the encoder and all three task specific decoders and the features are concatenated.} \label{fig:TopNet}
\end{figure}
\begin{figure}[]
\centering
\includegraphics[width=0.8\linewidth]{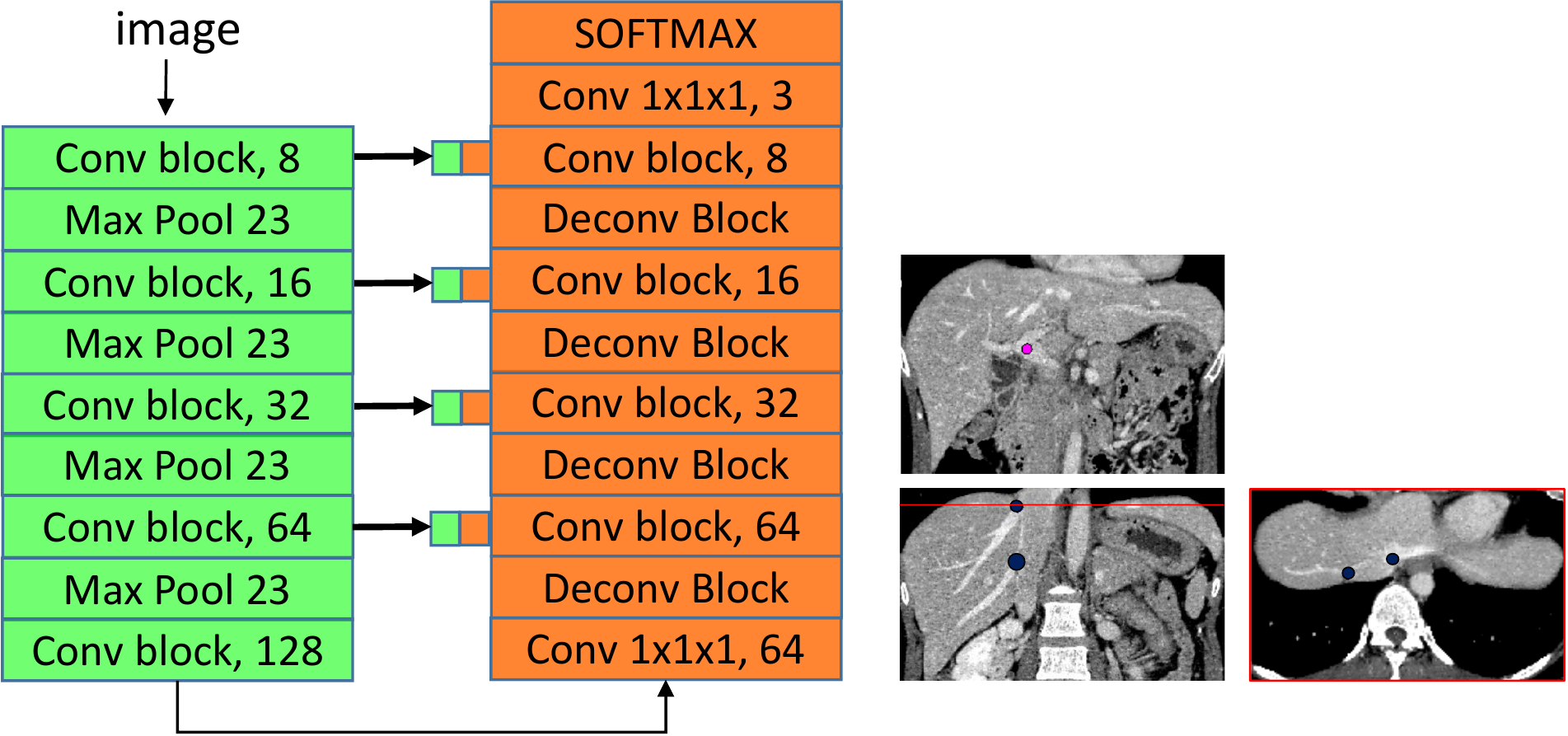}
\caption{VesselSource detector. Left: Network architecture which outputs a 3 channel probability map for background region, portal vein source and hepatic vein source points respectively, Right: detected portal vein (pink) and hepatic vein (blue) sources.} \label{fig:VesselSource detector}
\end{figure}
\begin{figure}[]
\centering
\includegraphics[width=\linewidth]{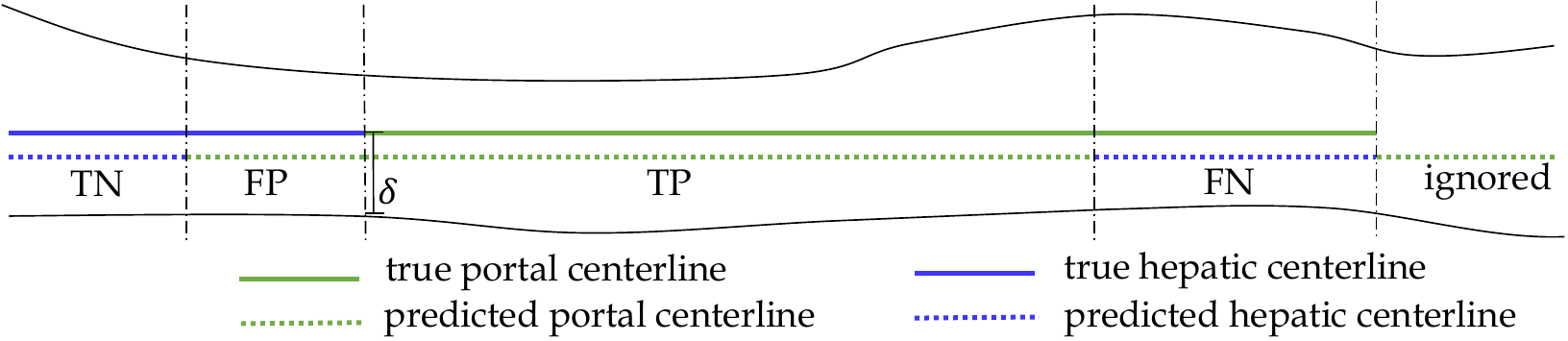}
\caption{An illustrations of terms used in the evaluation metrics.} \label{fig:Metric}
\end{figure}
\begin{figure}[]
\centering
\includegraphics[width=\linewidth]{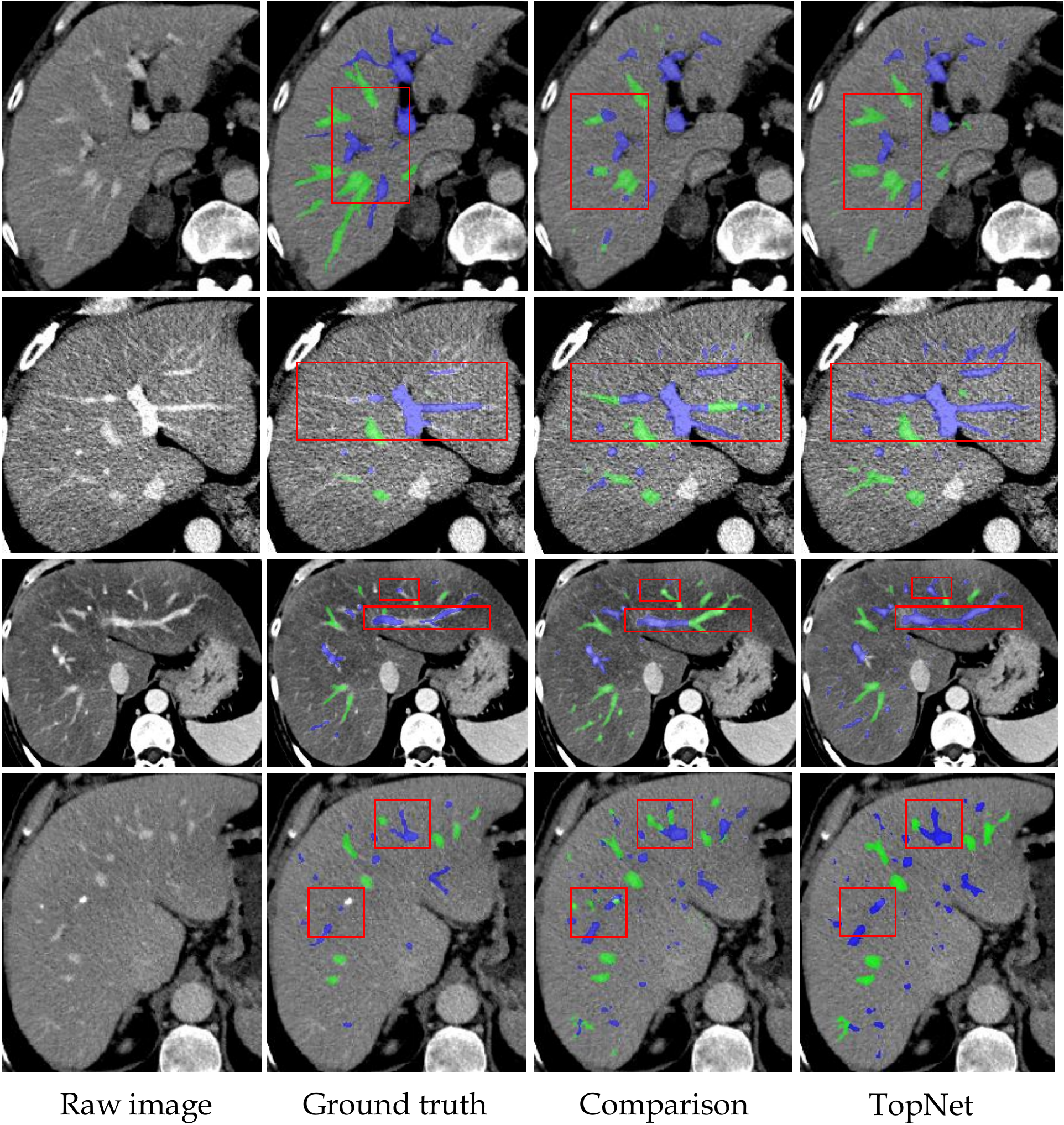}
\caption{Comparison of TopNet with different approaches on IRCAD datset. Red boxes show regions of interest. First two rows: Single-task 3D-FCN Vs TopNet. Third now: Multi-task Cosine metric Vs TopNet. Fourth row: Multi-task classification Vs TopNet.} \label{fig:Comparison1}
\end{figure}